\documentclass[a4paper]{article}

\usepackage{spconf,amsmath,graphicx}
\usepackage{CJKutf8}

\title{SNDCNN: Self-Normalizing Deep CNN\MakeLowercase{s} With \\ Scaled Exponential Linear Units For Speech Recognition}
\name{\begin{tabular}{c}Zhen Huang, Tim Ng, Leo Liu, Henry Mason, Xiaodan Zhuang, Daben Liu \end{tabular}}
\address{Apple Inc., USA \\
\texttt{\{zhen\_huang,tim\_ng,lliu9,hmason,xiaodan\_zhuang,daben\_liu\}@apple.com}} 
\begin{document}
 \ninept
\maketitle
\begin{abstract}
Very deep CNNs achieve state-of-the-art results in both computer vision and speech recognition, but are difficult to train. 
The most popular way to train very deep CNNs is to 
use shortcut connections (SC) together with batch normalization (BN).
Inspired by Self-Normalizing Neural Networks, we propose  the self-normalizing deep CNN (SNDCNN) based acoustic model topology, by removing the SC/BN and replacing the typical RELU activations with scaled exponential linear unit (SELU) in ResNet-50. 
SELU activations make the network self-normalizing and remove the need for both shortcut connections and batch normalization. 
Compared to ResNet-50, we can achieve the same or lower (up to 4.5\% relative) word error rate (WER) 
while boosting both training and inference speed by 60\%-80\%.
We also explore other model inference optimization schemes to further reduce latency for production use.

\end{abstract}
\noindent\textbf{Index Terms}: shortcut connection, batch normalization, scaled exponential linear units, self-normalization, ResNet, very deep CNNs

\section{Introduction}
\vspace{-3mm}

Very deep CNNs achieve state-of-the-art results on various tasks \cite{simonyan2014very} in computer vision.
Network depth has been crucial in obtaining those leading results \cite{simonyan2014very, szegedy2015going}. 
Naïve deep stacking of layers typically leads to a vanishing/exploding gradients problem, making convergence difficult or impossible.
For example, VGGNet \cite{simonyan2014very} only uses 18 layers.
Normalization methods, including batch normalization \cite{ioffe2015batch}, layer normalization \cite{ba2016layer} and weight normalization \cite{salimans2016weight}, allow deeper neural nets to be trained.
Unfortunately, these normalization methods make training stability sensitive to other factors, 
such as SGD, dropout, and the estimation of normalization parameters. Accuracy often saturates and degrades as network depth increases \cite{he2015convolutional, srivastava2015highway}.

ResNet \cite{He_2016_CVPR} uses shortcut connections (SC) and batch normalization (BN), 
allowing the training of surprisingly deep architectures with dramatic accuracy improvements. 
Since its invention, ResNet has dominated the field of computer vision.
The later state-of-the-art-model, DenseNet \cite{huang2017densely}, also uses SC and BN. 
Besides success in computer vision, 
ResNet has also performed well in acoustic models for speech recognition \cite{saon2015ibm, xiong2018microsoft}.

An alternative solution to the problem of vanishing/exploding gradients is self-normalizing neural networks\cite{klambauer2017self}. SNNs
use the scaled exponential linear unit (SELU) activation function to induce self-normalization. SNNs have been shown to converge very deep networks without shortcut connections or batch normalization. SNNs are also robust to perturbations caused by training regularization techniques.

Very deep convolutional neural network acoustic models are computationally expensive when used for speech recognition. Several techniques have been explored to improve inference speed on commodity server CPUs. Batching and lazy evaluation have been shown to improve inference speed on CPUs \cite{vanhoucke2011cpus} for neural networks of all types. Specifically for speech recognition, running inference at a decreased frame rate \cite{vanhoucke2013frames} has also been shown to reduce computation cost without affecting accuracy too much. We use frame-skipping and multi-threaded lazy computation.

Inspired by \cite{klambauer2017self}, we propose another way to train very deep networks without SC and BN by utilizing SELU activations.		
Experimental results in speech recognition tasks show that 		
by removing the SC/BN and replacing the RELU activations with SELU in ResNet50, 		
we can always get lower WER (up to 4.5\% relative) than ResNet50 		
and 60\%-80\% training and inference speedup. 		
We further speech up the decoding by applying techniques such as frame skipping and multi-thread lazy computation.

\begin{figure}[h]
	\centering
	\includegraphics[width=0.75\linewidth]{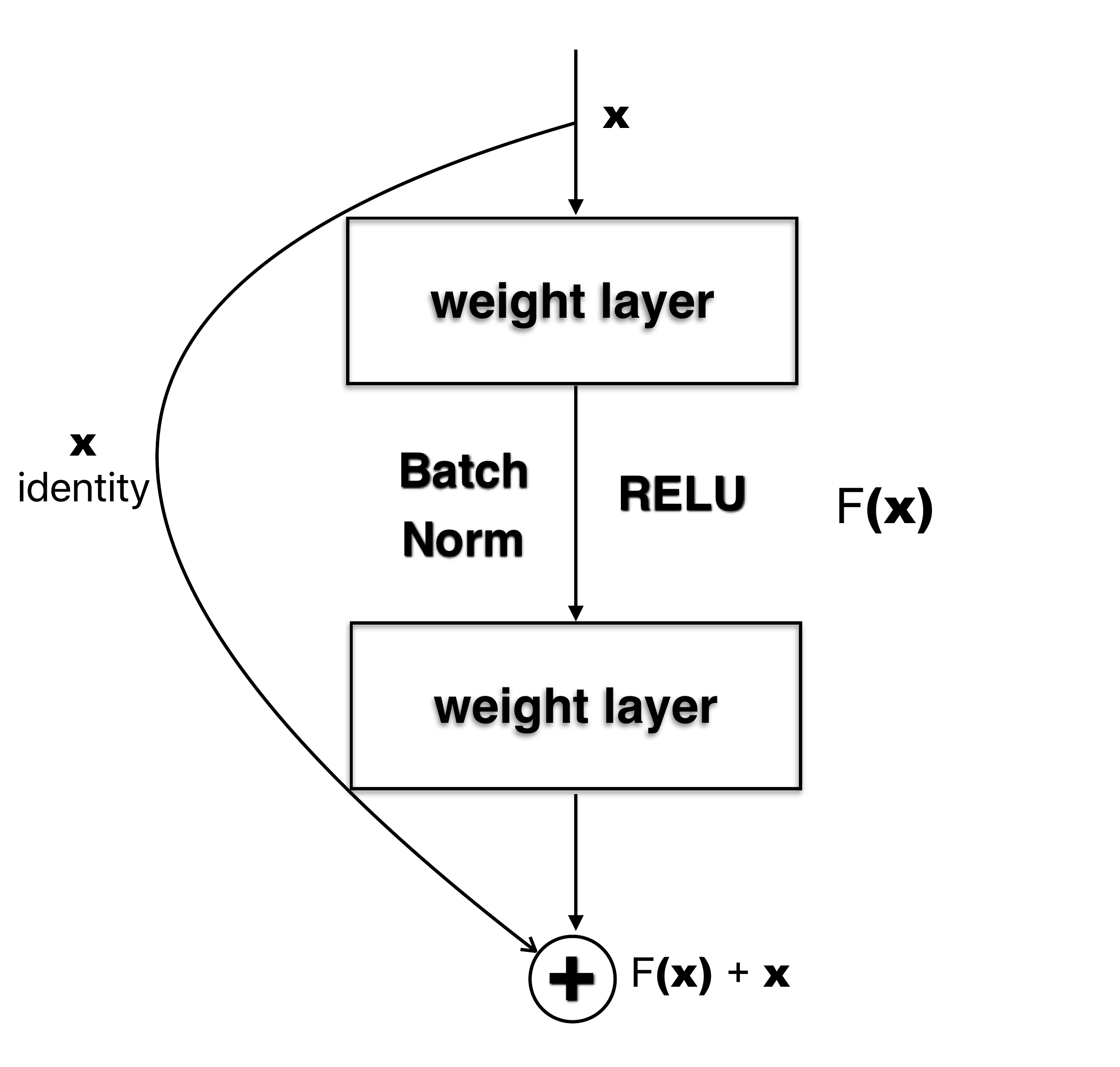}
	\caption{Typical building block of ResNet }
	\label{fig:resnet}
\end{figure}
\section{Related Work}
\label{sec:RL}
\subsection{Residual Learning}
\label{ssec:RL}
ResNet \cite{He_2016_CVPR} solves many problems in training very deep CNNs. 
The key ResNet innovation is the shortcut connections shown in Figure \ref{fig:resnet} which depicts a typical building block of ResNet. 
The input to the block, $x$, will go through both the original mapping $F(x)$ 
(weight layers, RELU activations and batch normalization \cite{ioffe2015batch}) and 
 the identity shortcut connection. 
The output, $y$, will be $F(x)+x$. 
The authors in \cite{He_2016_CVPR} hypothesize that the so-called residual mapping of $y=F(x)+x$ 
should be easier to optimize than the original mapping of $y=F(x)$. 
The design of the special building block is motivated by the observation in \cite{he2015convolutional, srivastava2015highway} 
that accuracy degrades when more layers are stacked onto an already very deep CNN model.  
If the added layers can be constructed as identity mappings, 
the deeper model should not have worse training error 
than the original shallower model without these added layers. 
The degradation actually suggests 
that the optimizer has difficulties in approximating identity mappings.  
With the identity shortcut connections in the ResNet building block, 
the optimizer can simply drive the layer weights toward zero 
to make the block identity mapping.
ResNet-style CNNs have maintained state-of-the-art results and have inspired other model structures \cite{huang2017densely,szegedy2017inception}.

\subsection{Batch Normalization}
\label{ssec:BN}
Besides the shortcut connections shown in Figure \ref{fig:resnet}, 
batch normalization (BN) \cite{ioffe2015batch} is also an important feature of ResNet. 
BN is designed to reduce internal covariate shift, defined 
as the change in the distribution of network activations 
due to the change in network parameters, during training.
This ensures better and faster convergence of the training process. 
BN is achieved by whitening the input of each layer, but full whitening of each layer’s inputs is costly
and not differentiable everywhere. Instead of whitening
the features in layer inputs and outputs jointly, 
each scalar feature is normalized independently to zero mean and unit variance. For a layer
with d-dimensional input $x = (x(1) . . . x(d))$, each dimension will be normalized as:
 \begin{equation}
\widehat{x}^{(k)} = \frac{x^{(k)}-\mathbf{E}[x^{(k)}] }{\sqrt{\mathbf{Var}[x^{(k)}]}}
\label{eq:bn_1}
\end{equation}
BN also ensures that the normalization can represent the identity transform by introducing a pair of parameters $\gamma^{(k)}$, $\beta^{(k)}$, which scale and shift the normalized value $\widehat{x}^{(k)}$:
 \begin{equation}
y^{(k)} = \gamma^{(k)}\widehat{x}^{(k)} + \beta^{(k)}.
\label{eq:bn_2}
 \end{equation}
In mini-batch based stochastic optimization, 
the mean $\mathbf{E}[x^{(k)}]$ and variance $\mathbf{Var}[x^{(k)}]$ are estimated within each mini-batch. 

BN has been successfully adopted in various tasks, but 
training with BN can be perturbed by many factors 
such as SGD, dropout, and the estimation of normalization parameters.
Moreover, in order to fully utilize BN, 
samples in each mini-batch must be i.i.d \cite{ioffe2017batch}. 
However, state-of-the-art speech recognition requires sequence level training of the acoustic model \cite{vesely2013sequence}. 
In sequence level training, a mini-batch consists of all the frames of a single utterance, 
and the frames are highly correlated to each other. This violates the i.i.d requirement of BN, making batch normalization very challenging to use with sequence training.

\subsection{Self-Normalizing Neural Networks}
\label{ssec:SNNs}

\cite{klambauer2017self} introduces self-normalizing neural networks (SNNs) 
in which neuron activations automatically converge towards zero mean and unit variance. 
The key to inducing the self-normalizing properties in SNNs 
is the special activation function, the scaled exponential linear unit (SELU), formulated as:

\begin{equation}
\mathbf{selu}(x)=\lambda\left\{
\begin{array}{lr}
x &  \mathrm{if} \  x > 0\\
\alpha e^x - \alpha & \mathrm{if} \  x\leq 0
\end{array}
\right.
 \end{equation}
with $\alpha \approx 1.6733$ and $\lambda \approx 1.0507$.

\begin{figure}[h]
	\centering
	\includegraphics[width=0.75\linewidth]{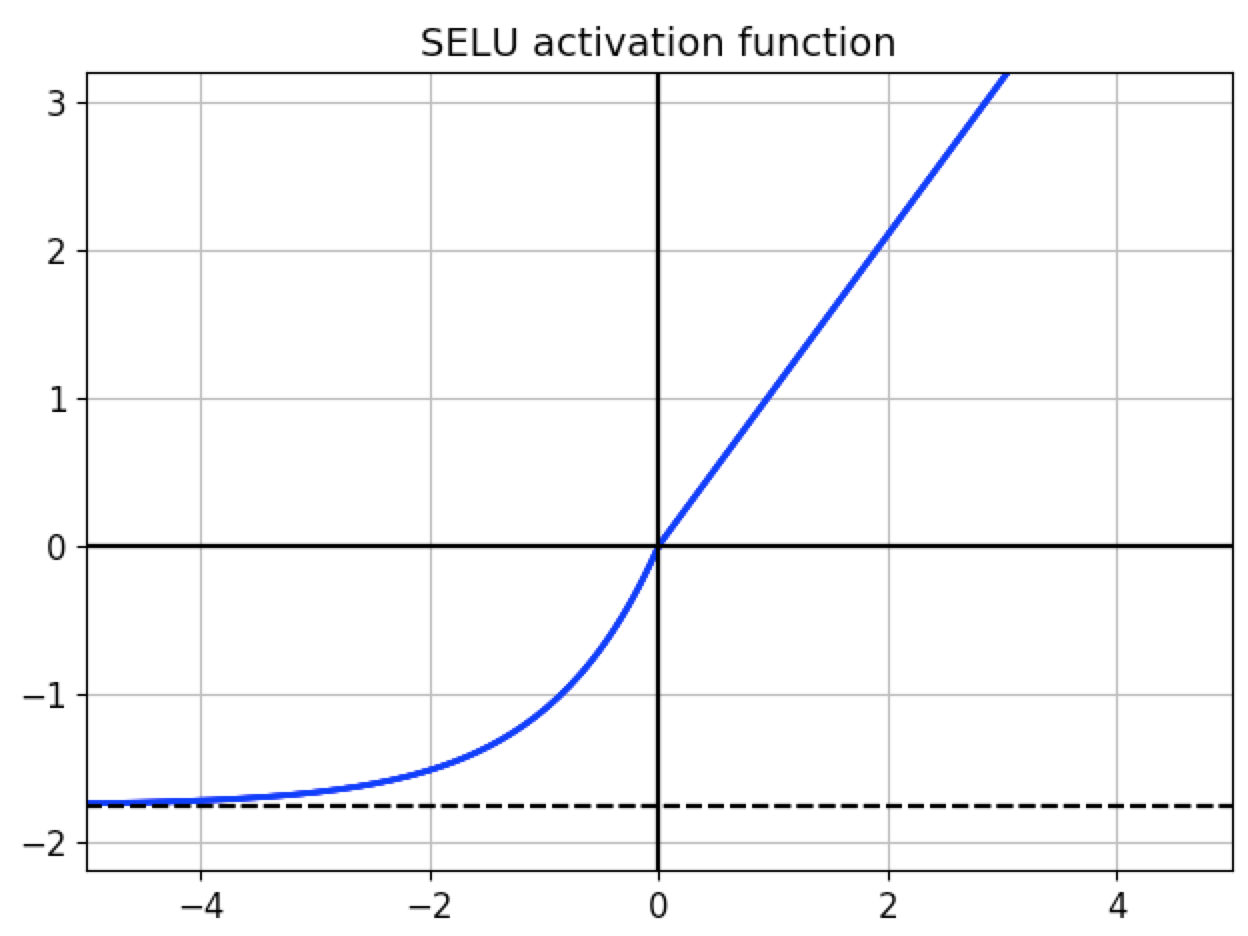}
	\caption{SELU activation function}
	\label{fig:selu}
\end{figure}

The values of $\alpha$ and $\lambda$ are obtained by solving fixed point equations to give the activation function the following characteristics, which ensures the self-normalizing property \cite{klambauer2017self}:

\begin{itemize}
\item[1] Negative and positive values for controlling the mean
\item[2] Saturation regions (derivatives approaching zero) to dampen the variance if it is too large in the lower layer
\item[3] A slope larger than one to increase the variance if it is too small in the lower layer
\item[4] A continuous curve
\end{itemize}

The shape of SELU activation function is shown in Figure \ref{fig:selu}. Using SELU, 
SNNs push neuron activations to zero mean and unit variance.
This gives us the same effect as batch normalization without being prone to the perturbations discussed in Section \ref{ssec:BN}.

\section{Training Self-Normalizing very deep CNNs}
\label{sec:}

\begin{figure}[h]
	\centering
	\includegraphics[width=0.75\linewidth]{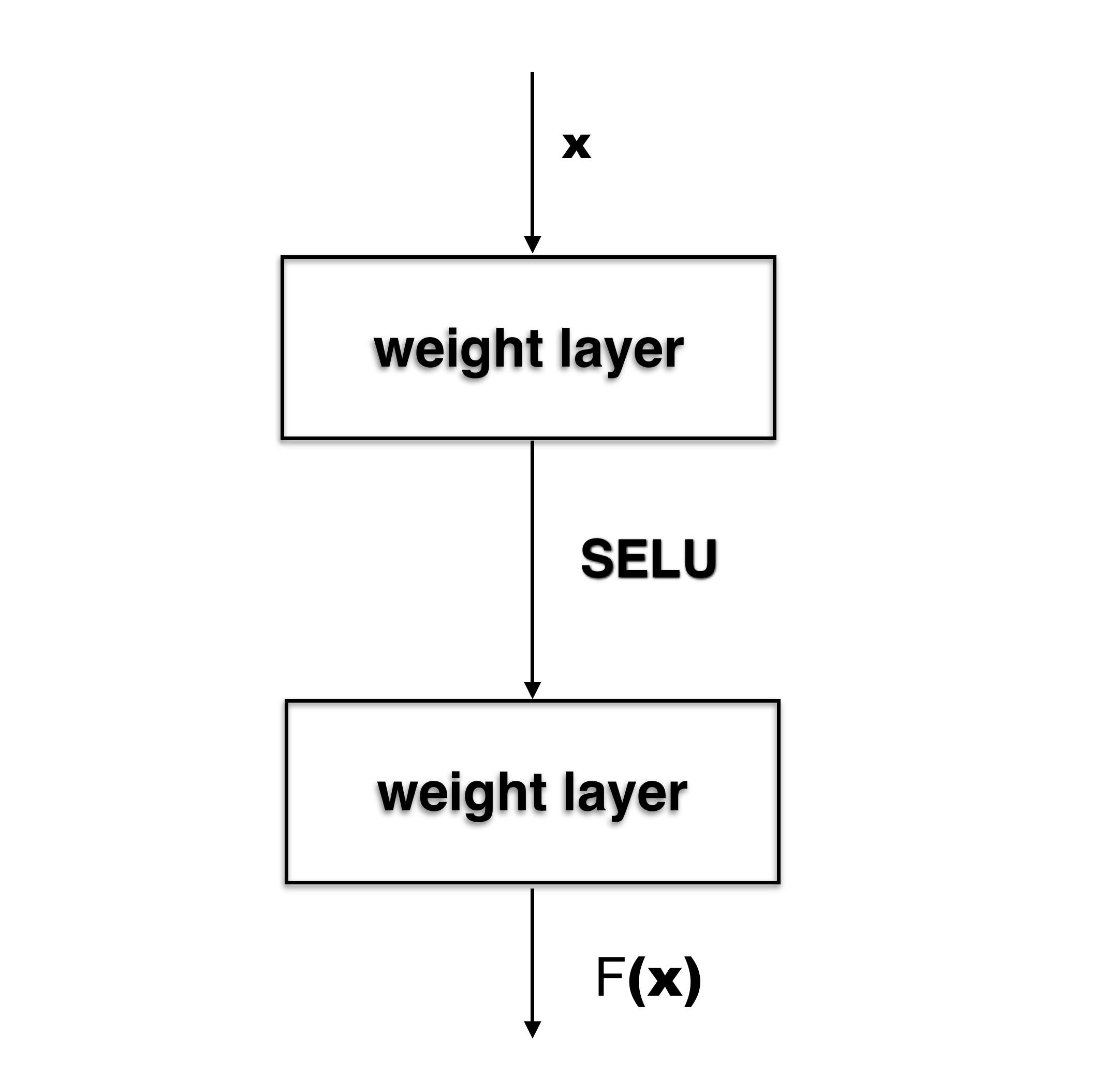}
	\caption{Building block of SNDCNN}
	\label{fig:rfdnn}
\end{figure}

We revise the model topology discussed in \cite{He_2016_CVPR} and design the proposed Self-Normalizing Deep CNNs (SNDCNN) for a hybrid automatic speech recognition system \cite{bourlard2012connectionist}.
The building block for SNDCNN is shown in Figure \ref{fig:rfdnn}. 
Comparing Figure \ref{fig:resnet} and \ref{fig:rfdnn}, we can see that the shortcuts and batch normalization are removed, and the activation function is changed to SELU. We thus practically obtain a self-normalizing ResNet.

\begin{figure}[h]
	\centering
	\includegraphics[width=\linewidth]{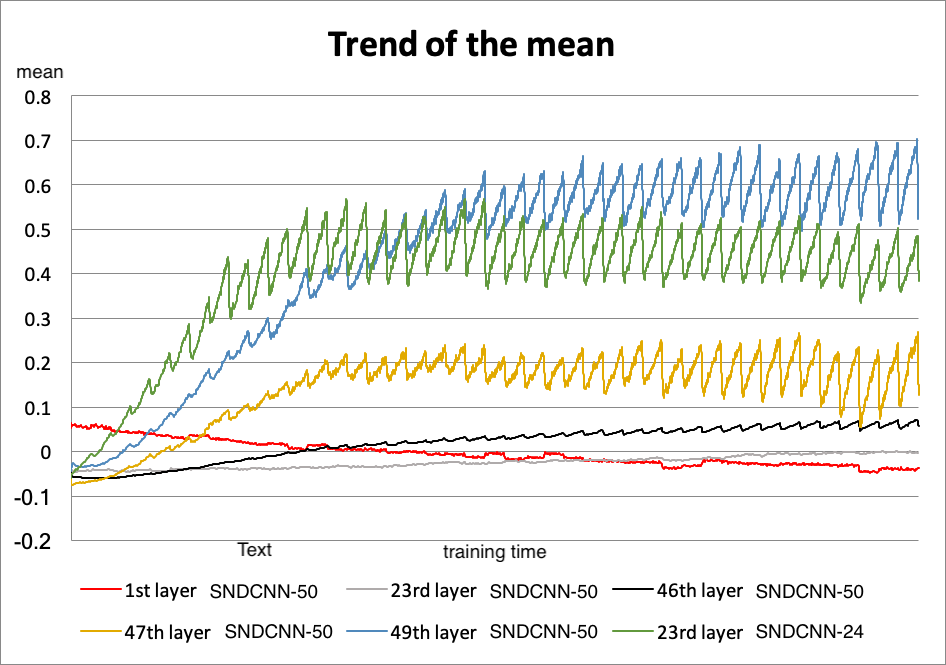}
	\caption{Trend of the mean}
	\label{fig:mean}
\end{figure}

\begin{figure}[h]
	\centering
	\includegraphics[width=\linewidth]{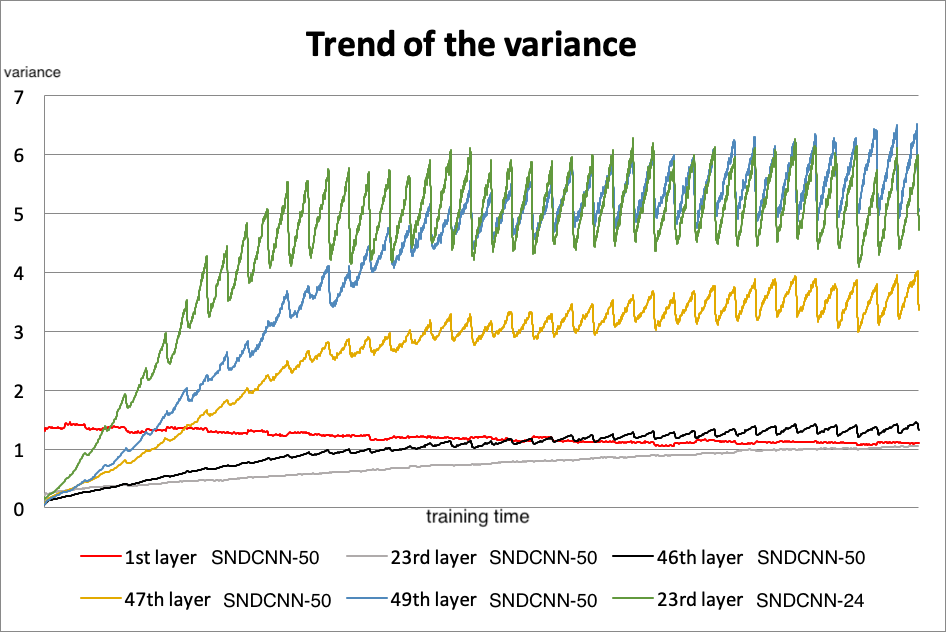}
	\caption{Trend of the variance}
	\label{fig:var}
\end{figure}

We verify the Self-Normalizing property by observing 
the trend of mean and variance in the SELU activation outputs during training. 
The model topology is a 50-layer CNN obtained by removing SC and BN from ResNet-50. We call this topology SNDCNN-50. 
Model parameters are initialized as instructed in \cite{klambauer2017self}. 
In Figures \ref{fig:mean} and \ref{fig:var}, we plot the mean and variance trend of the 1st, 23rd, 46th, 47th, and 49th layers of SNDCNN-50 and the 23rd layer of SNDCNN-24. 
The mean and variance are computed across frames within a mini-batch (256 frames). 
Each data point is obtained by averaging all the units in the same layer. 
The x-axis is training time, and we collect statistics from 33k mini-batches to draw each curve.

In the SNDCNN-50 case, we can see that the outputs of 1st and middle (23rd) layers follow the claims in \cite{klambauer2017self} nicely, but the last several layers do not. 
We find that the non-self-normalizing phenomenon becomes significant only after the 46th layer. As shown in Figure \ref{fig:mean} and \ref{fig:var}, the 46th layer almost has mean $=$ 0 and variance $=$ 1, but the following layers are worse.
We verify that the non-self-normalizing phenomenon is not caused by the depth of the neural network but by the distance to the output layer. 
The 23rd layer of SNDCNN-24 has the non-self-normalizing phenomenon, similar to the one seen in the 49th layer of SNDCNN-50, 
while the 23rd layer of SNDCNN-50 has a very nice self-normalizing property. 
We suspect that the back propagation path has to be long enough to effectively train the neural network's parameters to ensure the self-normalizing property.
Although the last layers do not strictly follow \cite{klambauer2017self}'s self-normalizing claim, 
the mean and variance are reasonable (mean $<$ 0.8, variance $<$ 9) even after 109 million mini-batches (28 billion training samples). 
We also evaluated different kinds of initialization for the network. 
Our findings indicate that as long as training starts normally, 
the trend of the mean and variance will follow the patterns seen in Figures \ref{fig:mean} and \ref{fig:var}.

Removing SC and BN simplifies the model structure and speeds up both training and inference. 
Removing BN also solves the sequence level training problem discussed in Section \ref{ssec:BN}. 
Most importantly, we always observe as good or better accuracy with the proposed simplified model structure.

\section{Experiments}
\label{sec:exp}
All  data used in this work comes from Siri internal datasets (en\_US and zh\_CN). All models are trained with Blockwise Model-Update Filtering (BMUF) \cite{chen2016scalable} with 32 GPUs. Newbob learning scheduling is used for all the experiments. A 4-gram language model is used in all experiments. 40 dimensional filter bank feature is extracted with 25ms window and 10ms step size. All the models use a context window of 41 frames (20-1-20) as the visible states \cite{mohamed2009deep}. 

\subsection{Accuracy}
\label{sec:acc}
Table \ref{exp:wer1} compares WERs of different model topologies for en\_US. 
The training data contains 300 hours of speech, and the testing data covers 7 hours of speech. 
From Table \ref{exp:wer1}, we have the following observations:
\begin{itemize}
	\item[1] [Row 1-4 vs. Row 5-8] Deep CNN models show advantage in terms of WER against shallower DNNs
	\item[2] [Row 3 vs. Row 4] [Row 7 vs. Row 8] SELU activation makes the training of very deep models (with no SC\&BN) feasible
	\item[3] [Row 1 vs. Row 2] [Row 5 vs. Row 6] SELU activation is no worse than RELU in DNN or ResNet topology.
	\item[4] [Row5 vs. Row 8] SNDCNN obtains better WER than ResNet
\end{itemize}
\begin{table}[t]\footnotesize
	\begin{center}
		\caption{WERs (in \%) of different model topologies with 300h training and 7h testing data in en\_US}
		\label{exp:wer1}
		\centerline{
			\begin{tabular}{|c|c|c|}
				\hline
				0&{\bf Model} & {\bf WER}  \\
				\hline
				1&6 layer DNN w/ RELU& 16.2\% \\
				\hline
				2&6 layer DNN w/ SELU&  16.0\%  \\
				\hline
				3&30 layer DNN w/ RELU &  not trainable\\
				\hline
				4&30 layer DNN w/ SELU &  15.9\%\\
				\hline
				5&ResNet-50 w/RELU w/ SC\&BN (standard ResNet) &  15.3\% \\
				\hline
				6&ResNet-50 w/SELU w/ SC\&BN &  15.2\% \\
				\hline
				7&ResNet-50 w/RELU w/o SC\&BN &  not trainable \\
				\hline
				8&ResNet-50 w/SELU w/o SC\&BN (SNDCNN-50) &  14.9\% \\
				\hline
		\end{tabular}}
	\end{center}
\end{table}	
\begin{table}[t]\footnotesize
	\begin{center}
		\caption{CERs (in \%) of different model topologies with 4000h training and 30h testing data in zh\_CN}
		\label{exp:wer2}
		\centerline{
			\begin{tabular}{|c|c|c|}
				\hline
				0&{\bf Model} & {\bf WER}  \\
				\hline
				1&ResNet-50 w/RELU w/ SC\&BN (standard ResNet) & 8.8\% \\
				\hline
				2&ResNet-50 w/RELU w/o SC w/ BN &  8.9\%  \\
				\hline
				3&ResNet-50 w/RELU w SC w/o BN &  8.7\%\\
				\hline
				4&ResNet-50 w/RELU w/o SC\&BN &  not trainable \\
				\hline
				5&ResNet-50 w/SELU w/ SC\&BN  &  8.7\% \\
				\hline
				6&ResNet-50 w/SELU w/o SC\&BN (SNDCNN-50)  &   8.7\% \\
				\hline
		\end{tabular}}
	\end{center}
\end{table}	
\begin{table}[t]\footnotesize
	\begin{center}
		\caption{WERs (in \%) with different model topologies with 10000h training and 7h testing data in en\_US}
		\label{exp:wer3}
		\centerline{
			\begin{tabular}{|c|c|c|}
				\hline
				0&{\bf Model} & {\bf WER}  \\
				\hline
				1& ResNet-50 & 8.8\% \\	
				\hline
				2&SNDCNN-50 &   8.4\% \\
				\hline
		\end{tabular}}
	\end{center}
\end{table}	
\begin{table}[t]\footnotesize
	\begin{center}
		\caption{Speedups (in \%) with different model topologies against standard ResNet-50}
		\label{exp:su1}
		\centerline{
			\begin{tabular}{|c|c|c|c|}
				\hline
				0&{\bf Model} & {\bf Training} & {\bf Inference}  \\ 
				\hline
				1&ResNet-50 & 0\% & 0\%\\	
				\hline
				2&ResNet-50 w/RELU w/o SC w/ BN  &   19.4\% & 30.0\% \\
				\hline
				3&ResNet-50 w/RELU w SC w/o BN   &   34.6\% & 49.7\% \\
				\hline
				4&SNDCNN-50  &   57.8\% & 80.6\%   \\
				\hline
		\end{tabular}}
	\end{center}
\end{table}	
Table \ref{exp:wer2} compares character error rate (CER) of different model topologies for zh\_CN.
The training data contains 4000 hours of speech and the testing data consists of 30 hours of speech. 
From Table \ref{exp:wer2}, we find that in order to make the training of very deep CNNs feasible, 
we must use at least one of the following three techniques: batch normalization, shortcut connection, and SELU activation. 
The WERs of different topologies with the same depth are actually very similar. 
This phenomenon suggests that depth could be the key to better accuracy. 
The proposed SNDCNN has slightly better WER than ResNet.

Table \ref{exp:wer3} compares en\_US WER of ResNet-50 and SNDCNN-50 with 10000 hours of training data and 7 hours of testing data.  
In this experiment, the proposed SNDCNN has much better WER than ResNet.
\subsection{Speedup}
\label{sec:speed}
Table  \ref{exp:su1} shows the relative computation speedups (frames per second) of the variants considered in Table \ref{exp:wer2}. From Table \ref{exp:wer2}, we know that the 4 models in Table  \ref{exp:su1} have very similar WER. but from Table \ref{exp:su1}, we can find that removal of BN and SC results in significant speedup in both training and inference. The speedup (especially in inference) is very important in deploying SNDCNN-50 in production systems where minimising latency is essential. 

\section{Inference Performance Optimization}
\label{sec:opt}

\begin{figure}[t]
	\centering
	\includegraphics[width=0.75\linewidth]{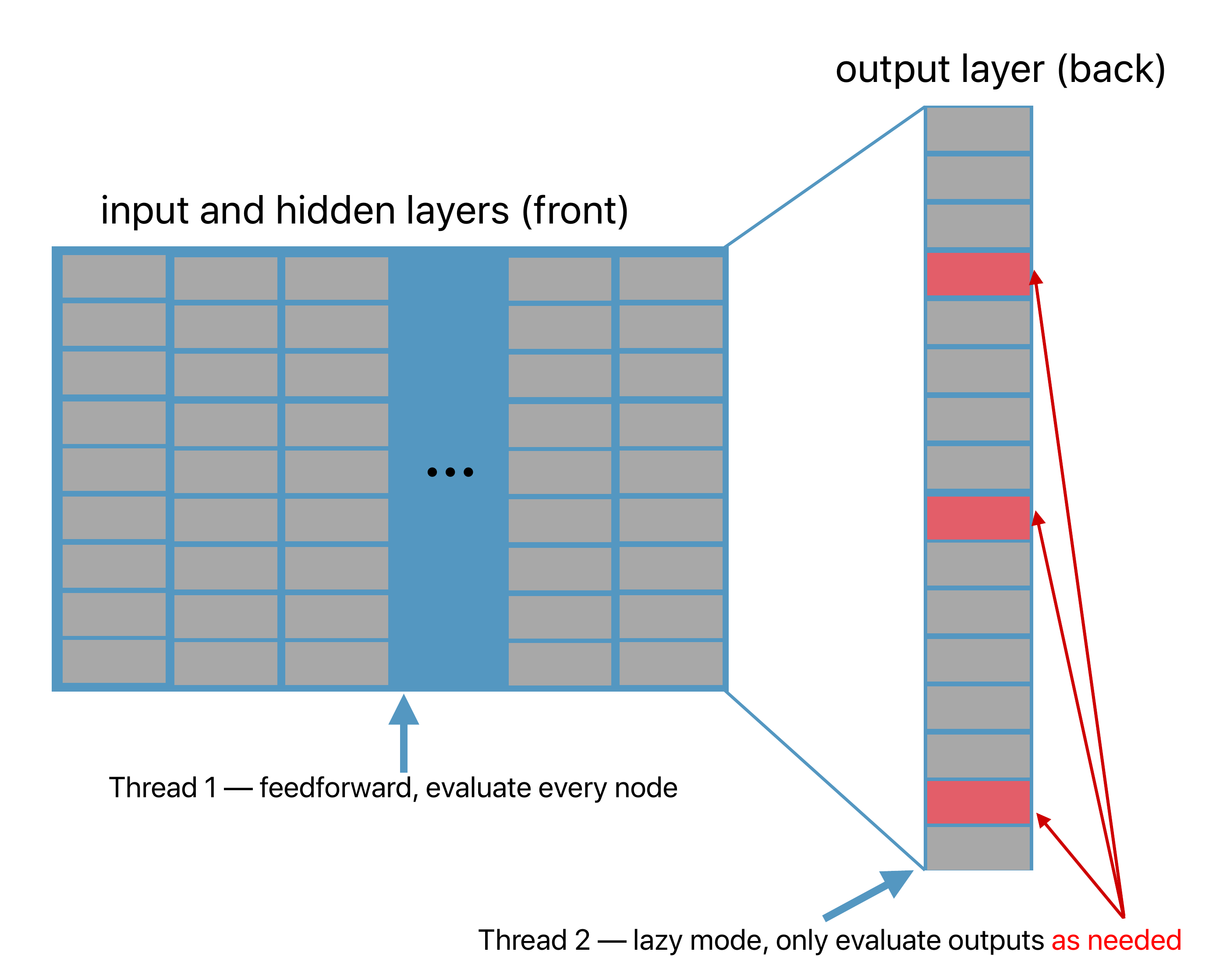}
	\caption{Multi-threaded lazy evaluation for acoustic model inference}
	\label{fig:lazy}
\end{figure}

We already achieve significant inference speedup by removing BN and SC from ResNet-50 as discussed in Section  \ref{sec:speed}. Further inference optimization for SNDCNN-50 was investigated, particularly frame-skipping and multi-threaded lazy computation.

\begin{table}[h]\footnotesize
	\begin{center}
		\caption{Latency reduction (in \%) with different inference techniques}
		\label{exp:su2}
		\centerline{
			\begin{tabular}{|c|c|c|}
				\hline
				0&{\bf Technique} & {\bf Latency reduction}  \\ 
				\hline
				1&Frame-skipping & 47.2\%\\	
				\hline
				2& Multi-thread lazy mode &   10.8\%  \\
				\hline
		\end{tabular}}
	\end{center}
\end{table}	
Frame-skipping \cite{vanhoucke2013frames}: our acoustic model targets tied HMM (hidden Markov model) states \cite{young1994tree}, running at 100 frames per second, but the predictions do not frequently change between frames. Human speech rarely has more than 10 phonemes per second. By simply skipping and duplicating two thirds of frames, we reduce the required computation by 3x which translates into 47.2\% latency reduction as shown in Table  \ref{exp:su2}. Note that usually skipping frames will result in some WER degradation \cite{vanhoucke2013frames} and we indeed observed that in our experiments with shallower models (10 layer, 2 convolution layer plus 8 fully connected) even when we  skip only half of the frames. However, with SNDCNN-50, we can skip up to two thirds of frames with no  degradation on WER.

Multi-thread lazy computation \cite{vanhoucke2011cpus}: as shown in Figure \ref{fig:lazy}, we split the acoustic model into two parts: front and back. We use two threads to do the inference independently. Thread 1 will do the inference of the front part which contains the input and hidden layers. Thread 2 will do the inference of the back part which contains the output layer. The outputs target
tied HMM states, and can easily be more than 10 thousand. As performing inference for the entire layer is expensive, we only compute the outputs that are needed by the decoding graph instead of computing every output of the layer. By doing this ``lazy" on-demand inference, we save a lot of computation in the large output layer, which translates into a 10.8\% latency reduction as shown in Table  \ref{exp:su2}.

\section{Conclusions}
\label{sec:con}
In this paper, we proposed a very deep CNN based acoustic model topology SNDCNN, by removing the SC/BN and replacing the typical RELU activations with scaled exponential linear unit (SELU) in ResNet-50. This leverages self-normalizing neural networks, by use of scaled exponential linear unit (SELU) activations, to train very deep convolution networks, instead of residual learning \cite{He_2016_CVPR}). With the self-normalization ability of the proposed network, we find that  the SC and BN are no longer needed. 
Experimental results in hybrid speech recognition tasks show that 
by removing the SC/BN and replacing the RELU activations with SELU in ResNet-50, 
we can achieve the same or lower WER
and 60\%-80\% training and inference speedup. 
Additional optimizations in inference, specifically frame skipping and lazy computation with multi-threading, further speed up the SNDCNN-50 model by up to 58\% which achieves production quality accuracy and latency.
\section{Acknowledgments}
The authors would like to thank Professor Steve Young, Bing Zhang, Roger Hsiao, Xiaoqiang Xiao, Chao Weng and  Professor Sabato Marco Siniscalchi for valuable discussions and help.
\newpage

\bibliographystyle{IEEEtran}
\bibliography{mybib}
\end{document}